\documentclass[conference]{IEEEtran}
\IEEEoverridecommandlockouts
\usepackage{cite}
\usepackage{amsmath,amssymb,amsfonts}
\usepackage{algorithmic}
\usepackage{graphicx}
\usepackage{textcomp}
\usepackage{xcolor}
\usepackage{url}
\def\BibTeX{{\rm B\kern-.05em{\sc i\kern-.025em b}\kern-.08em
    T\kern-.1667em\lower.7ex\hbox{E}\kern-.125emX}}
\begin{document}

\title{Product Quantization for Surface Soil Similarity \\
\thanks{Funded by the Intelligent Environmental Battlespace Awareness (IEBA) Army Program. \\
* Contact author}
}

 \author{\IEEEauthorblockN{ \underline{Haley Dozier}*}
\IEEEauthorblockA{\textit{Information Technology Laboratory} \\
\textit{U.S Army Engineer Research and Development Center}\\
3909 Halls Ferry Rd, Vicksburg, MS,  USA \\
Haley.R.Dozier@usace.army.mil}
\and
\IEEEauthorblockN{ Althea Henslee}
\IEEEauthorblockA{\textit{Information Technology Laboratory} \\
\textit{U.S Army Engineer Research and Development Center}\\
3909 Halls Ferry Rd, Vicksburg, MS,  USA \\
Althea.C.Henslee@usace.army.mil}
\and
\IEEEauthorblockN{ Ashley Abraham}
\IEEEauthorblockA{\textit{Information Technology Laboratory} \\
\textit{U.S Army Engineer Research and Development Center}\\
3909 Halls Ferry Rd, Vicksburg, MS,  USA \\
Ashley.N.Abraham@usace.army.mil}
\and
\IEEEauthorblockN{ Andrew Strelzoff}
\IEEEauthorblockA{\textit{Information Technology Laboratory} \\
\textit{U.S Army Engineer Research and Development Center}\\ 
3909 Halls Ferry Rd, Vicksburg, MS,  USA \\
Andrew.P.Strelzoff@usace.army.mil}
\and
\IEEEauthorblockN{Mark Chappell}
\IEEEauthorblockA{\textit{Environmental Laboratory} \\
\textit{U.S Army Engineer Research and Development Center}\\
3909 Halls Ferry Rd, Vicksburg, MS,  USA \\
Mark.A.Chappell@erdc.dren.mil}

}

\maketitle

\begin{abstract}

The use of machine learning (ML) techniques has allowed rapid advancements in many scientific and engineering fields. One of these problems is that of surface soil taxonomy, a research area previously hindered by the reliance on human-derived classifications, which are mostly dependent on dividing a dataset based on historical understandings of that data rather than data-driven, statistically observable similarities. Using a ML-based taxonomy allows soil researchers to move beyond the limitations of human visualization and create classifications of high-dimension datasets with a much higher level of specificity than possible with hand-drawn taxonomies. Furthermore, this pipeline allows for the possibility of producing both highly accurate and flexible soil taxonomies with classes built to fit a specific application. The machine learning pipeline outlined in this work combines product quantization with the systematic evaluation of parameters and output to get the best available results, rather than accepting sub-optimal results by using either default settings or best guess settings.

\end{abstract}


\section{Introduction}

Historically, efforts to develop soil classification systems to determine "types" of similar soil throughout the world depended upon expert selection of measures and boundaries among varying soil classes.  Although it is important to consider natural phenomenon in a systematic way, such systems include biases and artifacts of prior historical systems and largely lack any meaningful utility for data science and machine learning applications.  

For many machine learning pipelines, prior knowledge is almost always used. That is, subject matter experts and scientists typically include their prior knowledge of a subject area to down-select variables in the training and testing of a model. When properly performed, this a priori knowledge can increase model performance by reducing the dimensions of the data to address the curse of dimensionality and eliminate known collinearity within data sets. Additionally, subject matter experts are needed to assess the applicability of the model for addressing the questions the model is trying to answer. This fact is particularly the case for naturally occurring soils, where predicting soil “type” (i.e., morphological designation) and corresponding geochemical and physical behavior continue to frustrate model development.  Much of the error in soil biogeochemical modeling originates from uncertainties in the parameters describing the soil and sediment compartments \cite{RN2215, RN925}.  Attempts to overcome this problem procedurally through standardized experimental or analytical methods (such as via published ASTM, USEPA, or OECD methods) are challenged by our current inability to define thermodynamically the initial state of a soil system, as can be done in systems with homogenous solids – a challenge largely occurring due to the inherent heterogeneity of soils.  Inherent, multi-scaled, soil heterogeneity, existing geospatially throughout the earth, makes it theoretically impossible to represent any particular soil biogeochemical behavior with a single “universal” coefficient or simple geochemical dataset.  Without defining the soil’s context or class, it is impossible to fully appreciate the importance of the quantitative characteristics, and ultimately, discriminate one soil from another.

Our solution to this problem involves adding a soil "class" or "label" to unambiguously define a soil “type” or class of global surface soil samples.  While soil “type” has been recognized as essential for distinguishing a large variety of behaviors in different soils \cite{RN2104, RN2398, RN2228} \cite{RN2553, RN2408, RN803, RN801, RN2227, RN2415, RN2393, RN1744, RN802, RN2189}, no consistent application of a soil class has been reported in the scientific literature.  In our work, however, we endeavored to employ a soil morphological class, such as those contained in soil taxonomic systems - the most popular being the U.S. Department of Agriculture, Natural Resources Conservation Service (NRCS) Soil Taxonomy \cite{RN2459, RN2222}, and the World Reference Base or WRB \cite{RN2220}.  While the pedomorphological underpinnings of these designations make them highly discriminating \cite{RN2418, RN2621, RN2623}, these taxonomic systems are not utilized globally, with each country choosing to employ their own morphological class depending on the soil feature deemed important to the national interest.  Without an accurate, quantitatively comparable classification system, our ability to meaningful compare soils throughout the world, such as assess their similarity from one location to another will remain as a pervasive technological barrier.  Furthermore, the ability to compare and contrast predictions of soil behavior will remain elusive as well.  

 The objective for this work is to develop a new soil analog technology for comparing similar soils around the world. Here, the term, “analog”, refers to geospatial coordinates that statistically match the soil geochemical and geophysical characteristics to a particular geospatial point of interest. By using an unsupervised learning methodology called Product Quantization, we can reduce the computational expense of traditional similarity search methodologies by  decomposing the dataset’s features into lower dimensional spaces called subspaces, then quantizing and collapsing the reference set into a smaller collection of representative vectors, or centroids. The collection of centroids associated with the subspaces are represented with a specific combination of integers used to identify each data point with its appropriate cluster.  These centroids, or "codes", and the vectors associated with each subspace can be used as classes for similar groupings of soil and are held in an inverted index, which allows the future identification of approximate nearest neighbors. 

This paper will outline this unsupervised learning methodology for grouping similar soil using Product Quantization for specific purposes such as analog search and visualization.

\section{Related Work}

The application of machine learning methods to global soil mapping is still a relatively new development, and the results have been greatly debated in the soil field.  These machine learning based models, if accurate, are invaluable to researchers with fewer resources and less access to the immense range of soil observations necessary. However, there has been criticism as to the validity of both machine learning based models and the ability to precisely assess accuracy \cite{nature_meyer}. One of the most widely known models, Soil Grids \cite{SoilGrids} was refined by Poggio et al as Soil Grids 2.0 in 2020, using machine learning techniques combined with 240,000 global soil observations, performing well at coarse grain, but highlighting global scale spatial uncertainty. 

There are also those who offer a more reserved enthusiasm. As Meyer and Pebesma \cite{nature_meyer} recently noted, the results, though impressive at a surface level, have not withstood experts' opinions.  As soil data is carefully guarded by many countries, access is often limited, leading researchers to gather observations heavily from more easily accessible areas and sparsely from more restricted areas. This imbalance in the collection of data causes model accuracy to vary, with more sampled types being more accurately predictable.

\section{Data and Methodology}
In this section, we describe a methodology that uses product quantization that maps a given dataset to a smaller, fixed-dimensional dataset that can be used for clustering similar soil. Additionally, we propose a novel Regional LSH methodology that extends this methodology to develop soil analogs in regions of the world.  

\subsection{Data Description}

The soil dataset used for this study was obtained using an interpolated soil database published in 2017 by the International Soil Reference and Information Centre (ISRIC) \cite{SoilGrids}. This data set was interpolated using the raster package \cite{Raster} built for R programming language \cite{R_language}.  Here, individual geo-referenced rasters published by ISRIC containing various interpolated global soil characteristics were sampled at three different soil depths (0, 5, and 15 cm) against approximately 10 million randomly generated geospatial coordinates, and combined into a single dataframe. Afterwards, the data frame was preprocessed by removing rows with missing data (NANs) and negative values, and rows with duplicate geospatial coordinates.  Using the “over” command in the raster package, the dataset was further trimmed using a shapefile containing the global land mass boundaries to remove any rows containing erroneous geospatial coordinates.  The final dataset contained 6.7 million rows was then log-transformed (with the exception of the pH columns, in which were first transformed as $10^{-pH}$ before log transformation), mean centered, and normalized by the standard deviation of each column.  

\subsection{Product Quantization}

Quantization is a method that maps values from a large data set to values in a smaller, countable (finite) data set using a function known as a quantizer. This process can also be described as the mapping of a vector, $\textbf{x}\epsilon \mathbb{R}^D $, to a \textit{codeword} or \textit{centroid}, $\textbf{c}$, that is contained in a \textit{codebook} $C$. Traditional quantization techniques are non-linear in nature, and therefore do not preserve the ability to recover input values. An example of such a quantizer is the k-means algorithm,  which maps vectors $X=[\textbf{x}_1, \textbf{x}_2,...,\textbf{x}_N]$ from some Euclidean space of dimension $D$, $\mathbb{R}^D$, to a subset of itself with at most $k$ different values. The $k$ unique values are obtained by partitioning the data into $k$ mutually exclusive sets or clusters $S=[s_1, s_2, ..., s_k]$ such that the variance within the clusters is minimized. Often this is done through the optimization of some criterion function such as the Sum of Squared Error (SSE) function
\begin{equation}
    SSE = \sum_{i=1}^k \sum_{x_j \epsilon S_i} ||\textbf{x}_j-\textbf{c}_i||_2^2
\end{equation}
where $\textbf{c}_i$ is the centroid (or mean) of set $S_i$ and $||\cdot||_2 $ represents the $\ell_2$ (Euclidean) norm \cite{celebi2013comparative}. Once this optimization is complete, each vector $x_i$ can be mapped to the set $s_i$ corresponding to the minimum Euclidean distance between the set and vector means. This process allows the representation of each vector in the data set by the label of its centroid in an efficient manner.

Product Quantization (PQ) \cite{PQ, Optimized_PQ, nanopq} is a type of vector quantization method that can efficiently map, or encode, high dimensional vectors into a lower dimensional space. PQ is performed through the decomposition of the higher dimensional space into the Cartesian product, $C$, of low dimensional subspaces that can be quantized separately $C = C^1\times ... \times C^M$. That is, given a vector $\textbf{x} \ \epsilon \ \mathbb{R}^D $ that is the concatenation of $M$ subvectors $\textbf{x}=[\textbf{x}^1, \textbf{x}^2,...., \textbf{x}^M]$, a codebook of centroids can be created through the use of an objective function such as
\begin{equation}
    \min_{c^1,...,c^M} \sum_x ||x-c(i(x))||^{2}_{2} \label{objective_pq}
\end{equation}
such that $\textbf{c} \ \epsilon \  C = C^1\times ... \times C^M$ and $i(\textbf{x})$ represents an encoder that maps $x$ to the nearest codeword $\textbf{c}$  \cite{PQ,Optimized_PQ}.

\section{Point-based Similarity Search}

Once a collection of vectors $X$ is quantized, a similarity search can be performed by calculating the distance between a query vector and the vectors quantized using the process described in the previous section. If a given a query vector $\textbf{y} \ \epsilon \ \mathbb{R}^D $ is be represented by its centroid $i(y)$, then the distance $d(y,x)$ (often the Euclidean distance) between the query vector and vector $x \ \epsilon \ X$ can be approximated by
\begin{equation}
    d(\textbf{x},\textbf{y}) \approx d(i(\textbf{y}),i(\textbf{y})) = \sqrt{\sum_j d(i_j(\textbf{y}),i_j(\textbf{x}))^2} 
\end{equation}
where the distance between the $d(i_j(\textbf{y}),i_j(\textbf{x}))$ can be found using the look-up table created by the product quantizer that is associated with the $j^{th}$ quantizer. If the query vector $\textbf{y}$ is not encoded, then an asymmetric distance computation can be used where the squared distances are computed prior to the search \cite{PQ, RII}.






\section{Results}


To assess the results from using Product Quantization to encode the data in a lower dimensional space, we first examine two key performance parameters: computational efficiency and round-trip, or reconstruction, error. As previously stated, the data set used for this effort contains approximately 6.7 million rows with 48 features. 

\begin{figure}
    \centering
    \includegraphics[width = .5\textwidth]{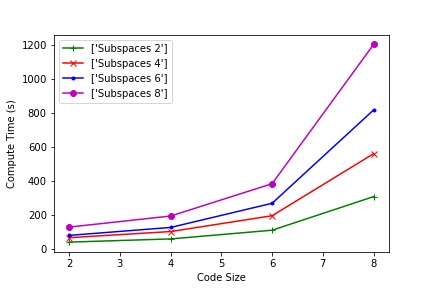}
    \includegraphics[width = .5\textwidth]{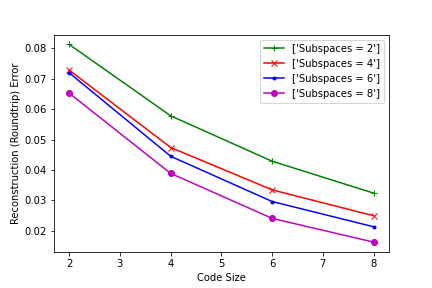}
    \caption{Important algorithmic parameters used for soil classification using the Python \cite{Python_language} based NanoPQ algorithm \cite{nanopq}. As code size and subspace number increase, computational expense (top) increases, but reconstruction error (bottom) decrease.}
    \label{fig:err_vs_time}
\end{figure}

Key parameters to this pipeline are subspace size and the number of centroids identified for each subspace. In practice, adaptable classification systems are desirable because each system can then be tailored for a specific application based on the needs of the user. This means that from a single dataset, there can be multiple classification systems for differing purposes, allowing morphological classes of surface soil that are adaptable for the specific decision-making need. Subspace size controls the trade-off between the model accuracy and memory-cost. A larger subspace size means higher accuracy (i.e. lower reconstruction error), but with higher memory-cost. The smaller the subspace size, the lower the memory usage will be, but also with lower accuracy (i.e. higher reconstruction error). Choosing the right subspace size will produce a balanced model.  Figure \ref{fig:err_vs_time} shows the trade off between parameters for the model.  
 
Choosing the appropriate clustering, or code, size balances the specificity of having more and smaller clusters with the usability of having fewer, but larger, clusters. To determine the best clustering size, an error estimate based on the round-trip error is analyzed relative to the number of classes. To determine the optimal amount of subspaces and clusters to use, a combinatorial search using these parameters can be completed to determine potential solutions for these objectives.

\begin{figure}[!ht]
    \centering
    \includegraphics[width = .5\textwidth]{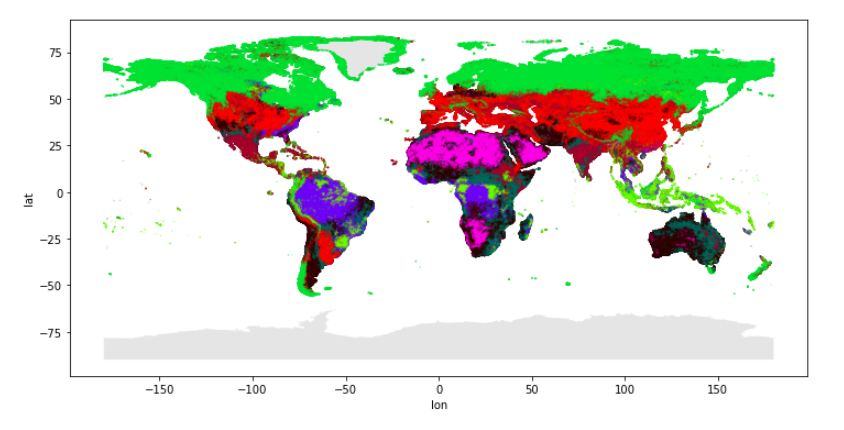}
    \includegraphics[width = .5\textwidth]{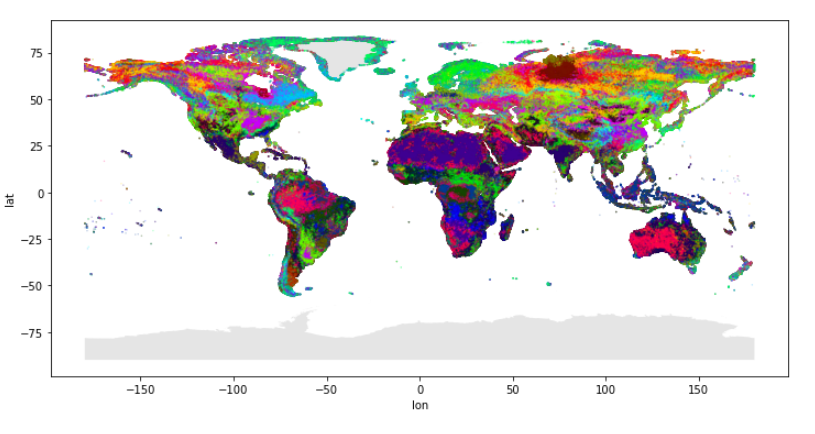}
    \caption{Global map colored by soil classes found using product quantization with 1 subspace, 32 centroids per subspace, and 32 possible soil classes (top) compared to 2 subspaces, 16 centroids per subspace, and 256 possible classes (bottom). This demonstrates the difference between the discernment of the model using different parameter choices. }
    \label{fig:map}
\end{figure}

Figure \ref{fig:map} shows the difference in the discrimination between two classification systems for surface soil created using product quantization with 1 subspace and 32 centroids per subspace  compared to 2 subspaces and 16 centroids per subspace. Using 1 subspace and 32 centroids per subspace allows for 32 possible classes of soil types within the world while  2 subspaces and 16 centroids per subspace allows for 256 possible classes. 

\begin{figure}
    \centering
    \includegraphics[width = .5\textwidth]{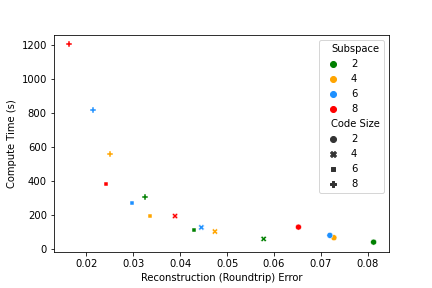}
    \includegraphics[width = .5\textwidth]{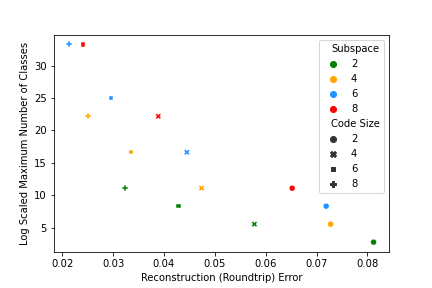}
    \caption{Trade-offs to consider during parameter selection in the model development process. As subspace and code size increase, the reconstruction error tends to decrease but computational expense tends to increase. }
    \label{fig:pareto}
\end{figure}


In practice, the determination of the number of soil classes can be determined through analyzing the trade-off between multiple parameters along with subject matter expert input. When a highly discriminating system is needed, we can select a model that will produce a high number of classes with low reconstruction error or compute time by examining Figure \ref{fig:pareto} and determining the best parameter selection based off necessary reconstruction accuracy and compute time.

\section{Conclusion and Future Work}

In this paper, we have introduced a method for detecting and grouping chemically similar regions of surface soil. By using this machine learning pipeline, we not only reduce the  computational expense of traditional similarity search methodologies, but also obtain a flexible classification system for developing soil classes, or types, that can be optimized to fit multiple application areas. 

\subsection{Future Work}

In the field of soil science, it is often the case that soil properties of one sample are too specific to identify in other parts of the world, and the number of soil data around from the same area identifies prominent properties and eliminates isolated random properties from individual soil data. Therefore, a significant future contribution to this field will include the extension of this work to a "regional" soil similarity search technology that can identify regional surface soils analogs throughout the world. Additionally, future work will address the hemispheric differences in the "types" of soil determined through the Product Quantization pipeline.


\bibliographystyle{splncs04}
%

\end{document}